\documentclass[letterpaper, 10 pt, conference]{ieeeconf}
    
\title{\LARGE \bf 
 CMetric: A Driving Behavior Measure using Centrality Functions}

\author{Rohan Chandra, Uttaran Bhattacharya, Trisha Mittal, Aniket Bera, and Dinesh Manocha \\
\small {University of Maryland, College Park}\\
Supplemental version including Tech Report, Code, Video, Datasets at \url{https://gamma.umd.edu/cmetric/}
\vspace{-12pt}
}

\newcommand{\mc}[1]{\mathcal{#1}}

\newcommand{\vts}[1]{\lvert #1 \rvert}
\newcommand{\Vts}[1]{\lVert #1 \rVert}
\newcommand{\bb}[1]{\mathbb{#1}}

\newcommand\Tstrut{\rule{0pt}{2.6ex}}         
\newcommand\Bstrut{\rule[-1.3ex]{0pt}{0pt}}   

\newcommand{\cm}{\mathcal{M}_{\Delta t}(u)}

\makeatletter
\newcommand\footnoteref[1]{\protected@xdef\@thefnmark{\ref{#1}}\@footnotemark}
\makeatother

\newcommand{\shorteq}{%
  \settowidth{\@tempdima}{-}
  \resizebox{\@tempdima}{\height}{=}%
}

\usepackage{amssymb,fge}

\usepackage{amsmath, amssymb}
\usepackage{subcaption}
\usepackage[utf8]{inputenc}
\usepackage[T1]{fontenc}
\usepackage[english]{babel}
\usepackage[linesnumbered,ruled,vlined]{algorithm2e}
\usepackage[font=small,belowskip=-1.5pt]{caption} 
\usepackage{array}
\usepackage{graphicx}
\usepackage{amsfonts}
\usepackage{soul}
\usepackage{hhline}
\usepackage{multirow, makecell}
\usepackage{float}
\usepackage{booktabs}

\usepackage{amsthm}
\usepackage{color}
\usepackage{transparent}
\usepackage{url}
\usepackage[symbol]{footmisc}
\usepackage{setspace}
\usepackage[colorlinks=true]{hyperref}
\usepackage{mathtools}

\DeclarePairedDelimiter\abs{\lvert}{\rvert}

\DeclareMathOperator*{\argmax}{arg\,max}
\DeclareMathOperator*{\argmin}{arg\,min}
\theoremstyle{plain}
\newtheorem{definition}{Definition}[section]

\newtheorem{problem}{Problem}[section]
\DeclareMathOperator{\EX}{\mathbb{E}}

\begin{document}

\maketitle
\thispagestyle{empty}
\pagestyle{empty}

\begin{abstract}
We present a new measure, CMetric, to classify driver behaviors using centrality functions. Our formulation combines concepts from computational graph theory and social traffic psychology to quantify and classify the behavior of human drivers. CMetric is used to compute the probability of a vehicle executing a driving style, as well as the intensity used to execute the style.  Our approach is designed for realtime autonomous driving applications, where the trajectory of each vehicle or road-agent is extracted from a video. We compute a dynamic geometric graph (DGG) based on the positions and proximity of the road-agents and centrality functions corresponding to closeness and degree. These functions are used to compute the CMetric based on style likelihood and style intensity estimates. Our approach is general and makes no assumption about traffic density, heterogeneity, or how driving behaviors change over time. We present an algorithm to compute CMetric and demonstrate its performance on real-world traffic datasets. To test the accuracy of CMetric, we introduce a new evaluation protocol (called ``Time Deviation Error'') that measures the difference between human prediction and the prediction made by CMetric. 


\end{abstract}
\section{Introduction}
\label{sec: introduction}

Autonomous driving is an active area of research with significant developments in perception, planning, and control, along with the integration of different methods and evaluation~\cite{schwarting2018planning}. Recent developments in perception technologies~\cite{chandra2019roadtrack,chandra2019densepeds} have resulted in good techniques for object recognition and tracking the positions of vehicles and road-agents using commodity visual sensors (e.g., cameras and lidars). One of the major challenges is to develop robust techniques for planning and decision-making, that can be used to compute collision-free and socially-acceptable trajectories for autonomous vehicles~\cite{chandra2019traphic, chandra2019robusttp}. This problem gets more challenging in urban environments, where the ego-vehicle is driving in close proximity to human drivers of vehicles, buses, trucks, bicycles, as well as pedestrians. A key issue in autonomous driving is safety, and the most important criteria is to compute safe and collision-free trajectories of the ego-vehicles. A recent trend is to classify and account for the driving behavior of other road agents and take them into account to perform behavior-aware planning~\cite{pnas, kuefler2017imitating, li2017game, pnas1}. 

There is considerable work in social psychology and traffic modeling on modeling and analyzing driver behaviors. Many studies from social traffic psychology~\cite{aljaafreh2012driving,ernestref16} conclude that driving behavior falls into three broad categories-- aggressive, neutral, and conservative. However, the exact definitions of these categories vary across the studies. Sagberg et al.~\cite{sagberg2015review} summarized these studies and developed a uniform definition such that each behavioral category can be determined in terms of specific \textit{styles} (See Table~\ref{tab: behaviors_centrality}). For example, aggressive driving may be manifested in styles such as overspeeding, overtaking, sudden lane-changes, etc. A style refers to a specific maneuver that a driver may perform and can be related to maneuver-based road-agent behavior~\cite{pseudobehavior2,honda}.

\begin{figure}[t]
\includegraphics[width = \columnwidth]{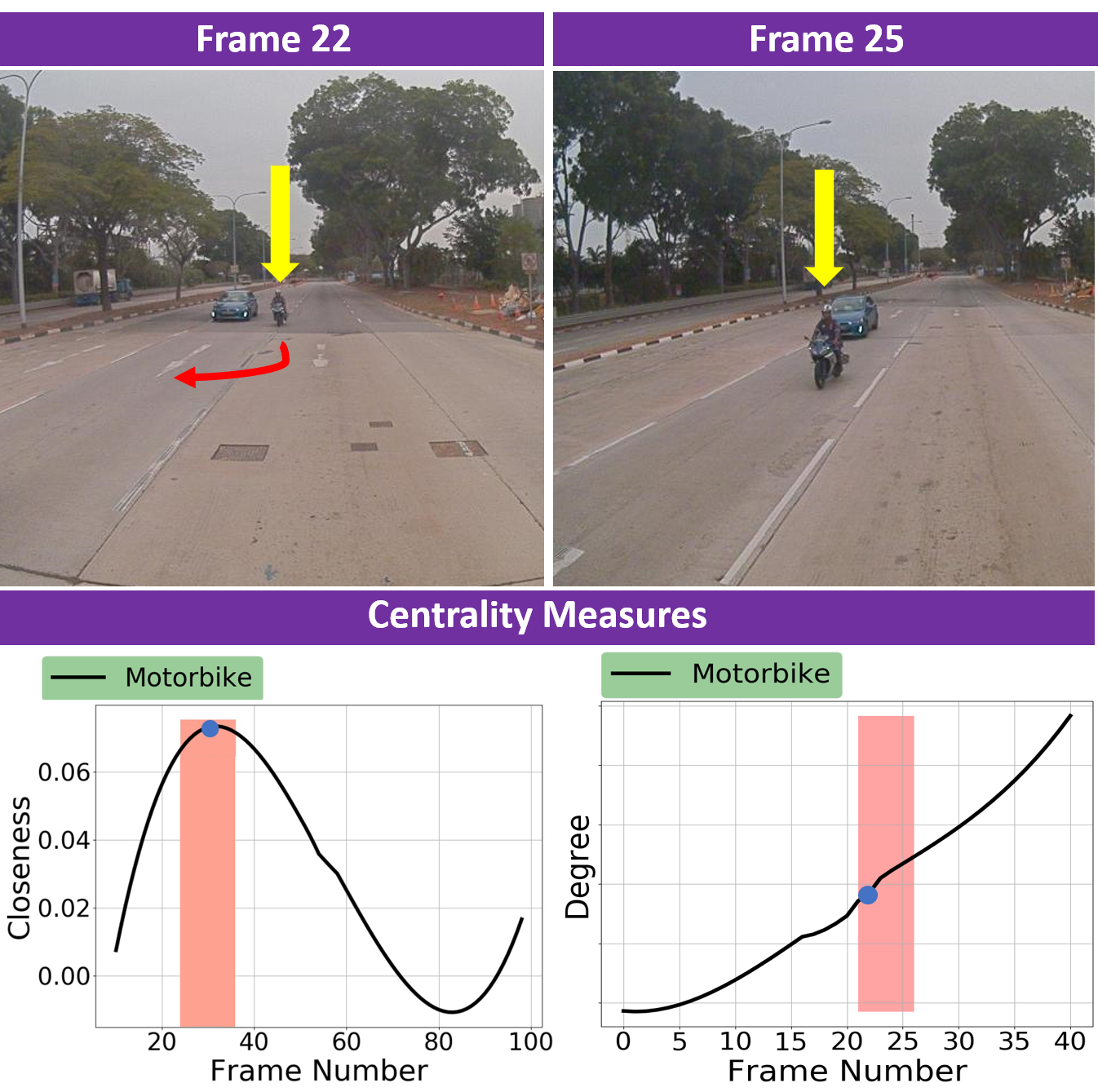}
\caption{\textbf{Modeling aggressive behavior using CMetric:} We use two centrality functions to measure different forms of aggressive styles such as weaving, overtaking, and overspeeding. CMetric characterizes overtaking by checking the critical points of the closeness centrality function, and models overspeeding by checking the rate of increase of the degree centrality function. In the example shown here, the bike is simultaneously overspeeding and overtaking the blue sedan between the $20^\textrm{th}$ and $30^\textrm{th}$ frames (red region). We therefore observe the critical point of the bike's closeness centrality around the $22^\textrm{nd}$ frame and the maximum slope of the degree centrality around the $25^\textrm{th}$ frame.}
\label{fig:1}
\vspace{-15pt}
\end{figure}
In this paper, we mainly focus on behavior prediction from realtime traffic videos. Some of the earlier work is based on data-driven or machine learning methods that rely on large datasets of traffic videos with behavior labels~\cite{honda,ernest}. While there are a lot of recent datasets for autonomous driving~\cite{honda, Argoverse,lyft2019}, they are mostly used for scene segmentation, object recognition, or vehicle trajectories, and do not contain behavior labels. Some other methods have been proposed for automatically classifying behaviors from trajectories based on spectral analysis~\cite{chandra2019graphrqi}, neural networks~\cite{chandra2019forecasting}, and game theory~\cite{pnas, pnas1}. Some of these methods assume that the driver behavior does not change over a long trajectory, while other methods are probabilistic and require offline generation of suitable priors.

\begin{figure*}[t]
\centering
\includegraphics[width=\textwidth]{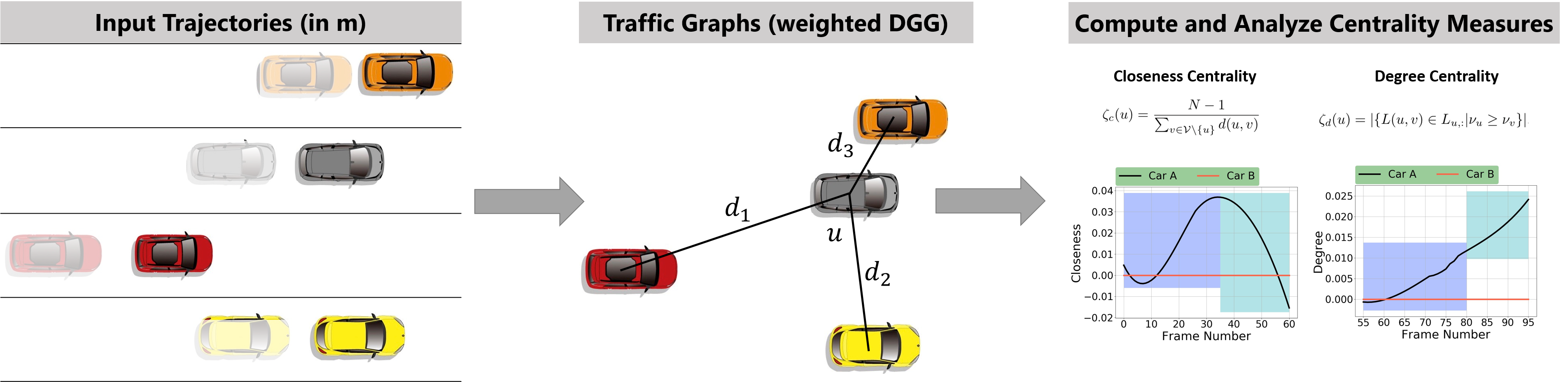}
\caption{\textbf{Overview of CMetric:} (\textit{left}:) Localization sensors on autonomous vehicle observe the positions of other vehicles; \textit{middle:} The positions and corresponding spatial distances between vehicles are represented through a weighted DGG (Section~\ref{subsec: DGG}); \textit{right:} Our CMetric uses the closeness and degree centrality functions to measure the style of each vehicle (Section~\ref{sec: centrality}). These styles are used to classify a global driving behavior (such as aggressive or conservative) for each vehicle (Table~\ref{tab: behaviors_centrality}).}
\label{fig: overview}
\vspace{-12pt}
\end{figure*}
The evaluation of behavior modeling methods is challenging due to the subjective nature of driver behavior. Standard evaluation protocols in the machine learning literature include the F1 score that measure the precision and recall, and have been successfully used in related fields such as human emotion recognition~\cite{mittal2020emoticon, mittal2019m3er}. In the case of driver behavior recognition, however, such protocols hold little meaning as a driver may exhibit both aggressive and conservative behavior depending on the context which changes rapidly with time. The goal then, is to classify the behavior with respect to context in a particular time interval.

\noindent\textbf{Main Contributions:} We present a realtime and general metric called CMetric for characterizing driving behavior based only on the positions of vehicles. We use graph theory to model the spatial interactions between the drivers through weighted dynamic geometric graphs. Our key insight for classifying driving behavior is based on distinguishing specific styles of drivers through vertex centrality measures~\cite{rodrigues2019network}. In graph theory and network analysis, centrality measures are real-valued functions on the vertices of a graph. We show that centrality functions have nice analytical properties that can be exploited to measure the likelihood and intensity of the driving styles. The CMetric measure is proposed as a combination of the centrality functions that we use to classify global behaviors and different styles.

We also propose a new evaluation protocol called Time Deviation Error (TDE), to evaluate driver behavior recognition methods. As a driver's behavior changes continuously in traffic, the goal must be to predict the correct behavior \textit{during the time-period in which that behavior is observed}. Our protocol measures the temporal difference between the time-stamp of an exhibited behavior in the ground-truth and time-period during which our model predicts the behavior. As an example, if a vehicle executes a rash overtake at the $5^\textrm{th}$ frame and our model predicts the maximum likelihood of the behavior at the $7^\textrm{th}$ frame, then the TDE$=0.067$ seconds assuming $30$ frames per second. A lower value for TDE indicates a more accurate behavior prediction model.

As compared to prior methods, our CMetric measure offers the following benefits: 1) It is a realtime algorithm that automatically operates on the data observed through cameras, and does not require any parameters to be adjusted manually and 2) CMetric can explicitly model the behaviors of surrounding human-driven vehicles in a deterministic manner.

\section{Related Work}
\label{sec: related work}
We give a brief overview of prior work in classifying driver behaviors, intent prediction, and graph-based traffic networks.
\subsection{Studies in Driver Behaviors}
\label{subsec: behavior_related}
At a broad level, the studies that analyze driver behaviors can be classified into three categories. 
The first category of analyses classify driver behavior based on the characteristics of drivers such as age, gender, blood pressure, personality, occupation, hearing, etc. Dahlen et al.~\cite{rohanref5} explored driver personalities and how they connected to aggressive driving. Rong et al.~\cite{rohanref3} investigated the causes of tailgating and determined that indicative features include blood pressure, hearing, and driving experience. Social Psychology studies~\cite{ernestref10} have found that aggressiveness may also be correlated to the background of the driver, including age, gender, occupation, etc. 

The second category of analyses is based on environmental factors such as weather or traffic conditions~\cite{behaviorref-category2-1,behaviorref-category2-2}. The study conducted in~\cite{behaviorref-category2-2} was designed to investigate the effects of weather-controlled speed limits and signs for slippery road conditions on driver behavior. Other studies~\cite{behaviorref-category2-1} have correlated changes in traffic density with varying driver behavior.

The final category of analyses refers to psychological aspects that affect driving styles. Psychological aspects include drunk driving, driving under influence, and state of fatigue. It is shown~\cite{behaviorref-category3-2} that driving under influence induces delayed responses in acceleration and deceleration. Jackson et al.~\cite{behaviorref-category3-3} show that a state of fatigue manifests the same characteristics as driving under the influence, but without the effect of substance intoxication. Other techniques evaluate the impact of mobile phone operation on driver behaviors~\cite{behaviorref-category3-1}. Our approach is complementary and can be combined with these methods.

\subsection{Behavior Prediction}
In contrast to offline driver behavior studies, many realtime algorithms have been proposed to learn a behavior model for human-driven vehicles. These methods are based on partially observable markov decision processes (POMDPs)~\cite{li2017game}, game theory~\cite{pnas, pnas1, sadigh2016information}, and imitation learning~\cite{kuefler2017imitating}. We refer the reader to Schwarting et al.~\cite{schwarting2018planning} for a more detailed review of these methods. The main disadvantage of these methods is that the behaviors of surrounding agents are modeled probabilistically, and thus exposing a high sensitivity to noise in the observed data. Probabilistic, or non-deterministic, approaches additionally rely on suitable prior distributions that may or may not be available. Our approach is deterministic and does not rely on a prior probability distribution.

\subsection{Graph-Based Traffic Networks}

Graph representations have been used to predict traffic flow~\cite{flow2} or traffic density~\cite{forecast1} at a macroscopic scale in applications such as congestion management and vehicle routing. Graphs have also been used for trajectory prediction~\cite{chandra2019forecasting,li2019grip,Yu2018SpatioTemporalGC} as well as action recognition~\cite{li2019learning}. The method proposed by Chandra et al.~\cite{chandra2019graphrqi} also models traffic entities using dynamic weighted graphs. However, they predict the driving behavior by training a neural network on the eigenvectors of the traffic-graphs. This approach requires a large amount of training data with behavior label annotations, which can be time-consuming and expensive to collect. Further, they assume that a driver's behavior is constant and does not change with time. Our centrality-based approach is more general and overcomes these limitations.


\section{Background and Overview}
\label{sec: background}
In this section, we give an overview of traffic driving behaviors, construction of traffic-graphs, and centrality functions. 

\subsection{Categorizing Driving Behavior}

Several criteria have been proposed in psychology and robotics literature to characterize driving behaviors. These include explicit formulations or scales to measure aggressive driving. The Driving Anger Scale (DAS)~\cite{deffenbacher1994development} consists of $14$ scenarios rated on a $5$-point Likert scale ($1=$not at all; $5=$very much) measuring the amount of anger experienced during an offensive situation (e.g.,  aggressive overtaking). The DAS assesses the propensity to become angry while driving and higher scores reflect greater driving anger. The DAS was extended to DAX (Driving Anger Expression) that identifies four ways people express their anger when driving, and they can be combined to form a Total Aggressive Expression Index.

More generally, other scales consider behaviors deduced from a question-answer based analysis. For example, the Multidimensional Driving Style Inventory (MDSI)~\cite{taubman2004multidimensional} developed a scale measuring eight factors, each one representing a specific driving style— dissociative, anxious, risky, angry, high-velocity, distress reduction, patient, and careful. Another scale, the Driving Behaviour Inventory (DBI), was developed to study dimensions of driver stress~\cite{gulian1989dimensions}, including driving aggression, dislike of driving, tension and frustration, and irritation. Similarly, the Driving Style Questionnaire (DSQ)~\cite{french1993decision} is composed of six independent dimensions of driving style that are labeled-- speed, calmness, social resistance, focus, planning, and deviance.

These scales measure global driving behaviors from different perspectives and therefore interpret the meaning of the behaviors in different ways that are hard to summarize. Instead, we follow the classification principle of Sagberg et al.~\cite{sagberg2015review}, which unifies these different interpretations to proposes that each global behavior is a function of specific driving maneuvers or \textit{styles}. This principle suggests that, rather than attempting to classify a vehicle's behavior according to global labels that are interpreted differently by different scales, it is more useful to predict the specific styles that constitute the global behavior. The principle proposed by~\cite{sagberg2015review} is as follows:
\begin{definition}
Driving behavior refers to the high-level \textit{global behavior}, such as aggressive or conservative driving. Each global behavior consists of one or more underling \textit{specific styles}. For example, an aggressive driver (global behavior) may frequently overspeed or overtake (specific styles). 
\label{def: behavior}
\end{definition}
We summarize the global behaviors and their constituent specific styles in Table~\ref{tab: behaviors_centrality}.
The scales described previously for measuring driving behavior cannot classify the specific styles in Table~\ref{tab: behaviors_centrality}. Our goal in this work is to develop a computational metric that measures the following specific styles-- \textit{Overtaking, overspeeding, sudden lane-changes, and weaving} from the trajectories. We state :

\begin{problem}
In a traffic video with $N$ vehicles during any time-period $\Delta t$, given the spatial coordinates in the world coordinate frame of all vehicles, our overall objective is to classify the \textit{specific styles} for all drivers during $\Delta t$ based on the styles described in Table~\ref{tab: behaviors_centrality}.
\end{problem}

\begin{table*}[t]
\centering
\caption{Definition and categorization of driving behaviors~\cite{sagberg2015review}. We measure the likelihood and intensity of specific styles by analyzing the first-and second-order derivatives of the centrality polynomials.}
\centering
\resizebox{\linewidth}{!}{
\begin{tabular}{lcccc}

\toprule

Global Behaviors & Specific Styles & Centrality & Style Likelihood Estimate & Style Intensity Estimate\\
\midrule
 \multirow{3}{*}{Aggressive}& Overspeeding & Degree ($\zeta_d$) & Magnitude of $1^\textrm{st}$ Derivative & Magnitude of $2^\textrm{nd}$ Derivative\\
 & Overtaking / Sudden Lane-Change & Closeness ($\zeta_c$) & Magnitude of $1^\textrm{st}$ Derivative  &   Magnitude of $2^\textrm{nd}$ Derivative\\
  & Weaving  & Closeness ($\zeta_c$) & Local Extreme Points & $\varepsilon$-sharpness of Local Extreme Points \\
 \midrule

\multirow{2}{*}{Conservative}& Driving Slowly or uniformly  & Degree ($\zeta_d$)& Magnitude of $1^\textrm{st}$ Derivative& Magnitude of $2^\textrm{nd}$ Derivative\\
 & No Lane-change  & Closeness ($\zeta_c$)& Magnitude of $1^\textrm{st}$ Derivative& Magnitude of $2^\textrm{nd}$ Derivative\\

\bottomrule
\end{tabular}
}
\label{tab: behaviors_centrality}
\end{table*}

\subsection{Traffic Representation Using Dynamic Geometric Graphs (DGGs)}
\label{subsec: DGG}

In this section, we describe the representation of traffic through Dynamic Geometric Graphs (DGGs).
We assume that the trajectories of all the vehicles in the video are extracted and given to our algorithm as an input.  Given this input, we first construct a DGG~\cite{waxman1988routing} at each time-step. We define a dynamic geometric graph as follows:

\begin{definition}
A Geometric Graph is an undirected graph with a set of vertices $\mathcal{V}$ and a set of edges $\mc{E} \subseteq \mc{V} \times \mc{V}$ defined in the 2-D Euclidean metric space with metric function $f(x,y) = \Vts{x-y}^2$. Two vertices $v_i, v_j \in \mc{V}$ are connected if, and only if, their $f(v_i,v_j) < r$ for some constant $r$.

A Dynamic Geometric Graph (DGG) is a geometric graph with a set of vertices $\mathcal{V}(t)$ and a set of edges $\mc{E}(t)$, where $\mc{V}(t)$ and $\mc{E}(t)$ are the sets of vertices and edges as functions of time.
\end{definition}

We represent traffic at each time instance with $N$ road-agents using a DGG, where the positions of vehicles, including motorbikes and scooters, represent the vertices. In particular, we represent a vehicle position as a point in $\mathbb{R}^2$. Thus, $v_i \gets [x_i,y_i]^\top$, where $ [x_i,y_i]^\top$ is the 2-D spatial coordinates (e.g., in meters) of the $i^\textrm{th}$ vehicle in the global coordinate frame.
For a DGG, $\mc{G}$, the adjacency matrix, $A \in \mathbb{R}^{N \times N}$ is given by,

\begin{equation}
A(i,j)=
     \begin{cases}
      d(v_i,v_j) & \text{if $d(v_i,v_j) < \mu,i \neq j$ },\\
      0 &\text{otherwise.}
     \end{cases}
     \label{eq: similarity_function}
\end{equation}
\noindent where $d(v_i,v_j)$ denotes the Euclidean distance between the $i^\textrm{th}$ and $j^\textrm{th}$ vehicles, and $\mu$ is a distance threshold parameter. 

For an adjacency matrix $A$ at each time instance, the corresponding degree matrix $D \in \mathbb{R}^{N \times N}$ is defined as a diagonal matrix with main diagonal $D(i, i) = \sum_{j=1}^N A(i, j)$ and 0 otherwise. Further, the symmetric Laplacian matrix can be obtained by subtracting $A$ from $D$, 

\begin{equation}
L(i,j) =
     \begin{cases}
       D(i,i)  &\text{if $i=j$},\\
      -d(v_i,v_j) &\text{if $d(v_i,v_j) < \mu$}, \\
      0 &\text{otherwise.}
     \end{cases}
\end{equation}
The Laplacian matrix for each time-step is correlated with the Laplacian matrices for all previous time-steps. Let the Laplacian matrix at a time instance $t$ be denoted as $L_t$. Then, the laplacian matrix for the next time-step, $L_{t+1}$ is given by the following update,

\begin{equation}
L_{t+1} =
\left[
\begin{array}{c|c}
L_{t} \Bstrut & 0 \Bstrut\\
\hline
0 \Tstrut & 1
\end{array}
\right] + \delta\delta^\top,
\label{eq: A_update}
\end{equation}

\noindent where $\delta \in \bb{R}^{(t+1) \times 2} $ is a sparse matrix with $\Vts{\delta}_0 \ll n$. The update rule in Equation~\ref{eq: A_update} enforces a vehicle to add edges connections to new vehicles while retaining edges with previously seen vehicles. The presence of a non-zero value in the $j^\textrm{th}$ row of $\delta$ indicates that the $j^\textrm{th}$ road-agent has formed an edge connection with a new vehicle, that has been added to the current DGG. The size of $L_t$ is fixed for all time $t$ and is initialized as a zero matrix of size $N$x$N$, where $N$ is max number of agents. $L_t$ is updated in-place with time and is reset to a zero matrix once the number of vehicles crosses $N$. The diagonal elements of the Laplacian matrix in Equation~\ref{eq: A_update} is used by the degree centrality to characterize overspeeding~\ref{sec: centrality}.

\subsection{Centrality Functions}
\label{subsec: background_centrality}


\begin{figure*}
\centering
  \begin{subfigure}[h]{\textwidth}
    \includegraphics[width=\textwidth]{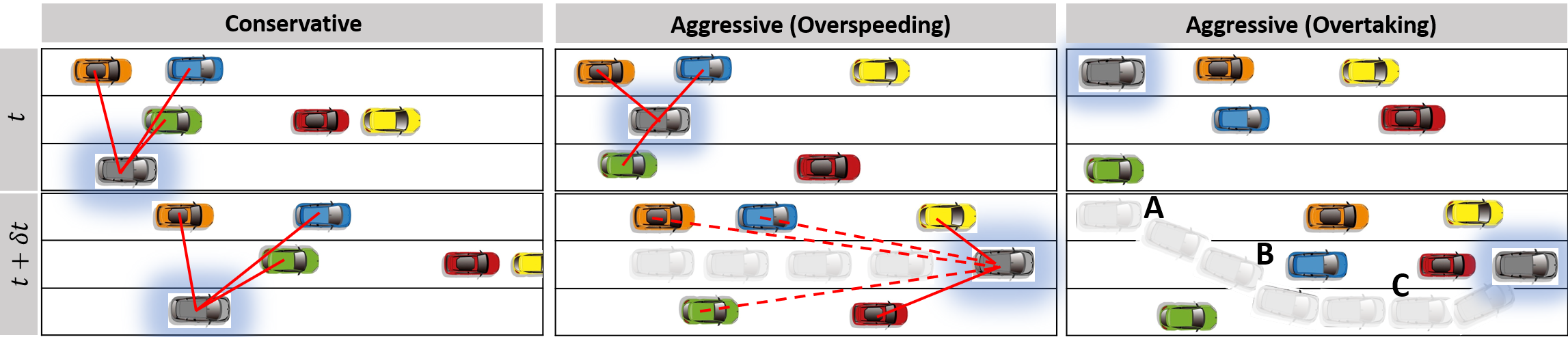}
    \caption{In all three scenariso, the ego-vehicle is a gray vehicle marked with blue glow outline. (\textit{left}) A conservative vehicle, \textit{(middle)} overspeeding vehicle in the same lane, and \textit{(right)} weaving and overtaking vehicle. }
    \label{fig: Frame 1}
  \end{subfigure}
  \begin{subfigure}[h]{0.328\textwidth}
    \includegraphics[width=\textwidth]{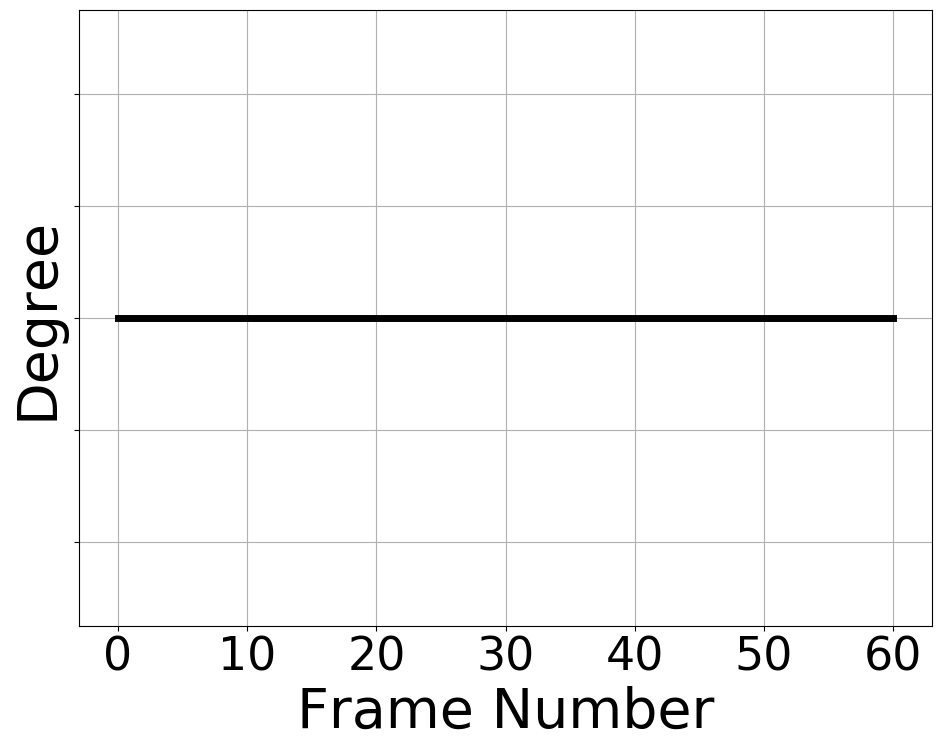}
    \caption{Constant degree centrality function for conservative vehicle.}
    \label{fig: Frame 1}
  \end{subfigure}
    \begin{subfigure}[h]{0.332\textwidth}
    \includegraphics[width=\textwidth]{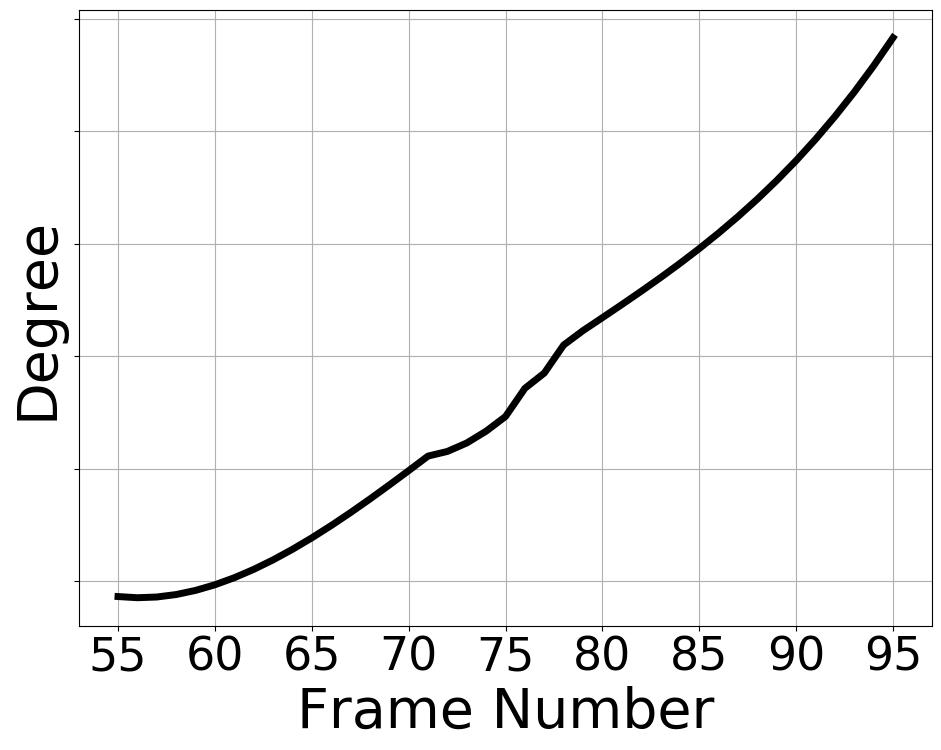}
    \caption{Monotonically increasing centrality function for overspeeding vehicle.}
    \label{fig: Frame 2}
  \end{subfigure}
  \begin{subfigure}[h]{0.324\textwidth}
    \includegraphics[width=\textwidth]{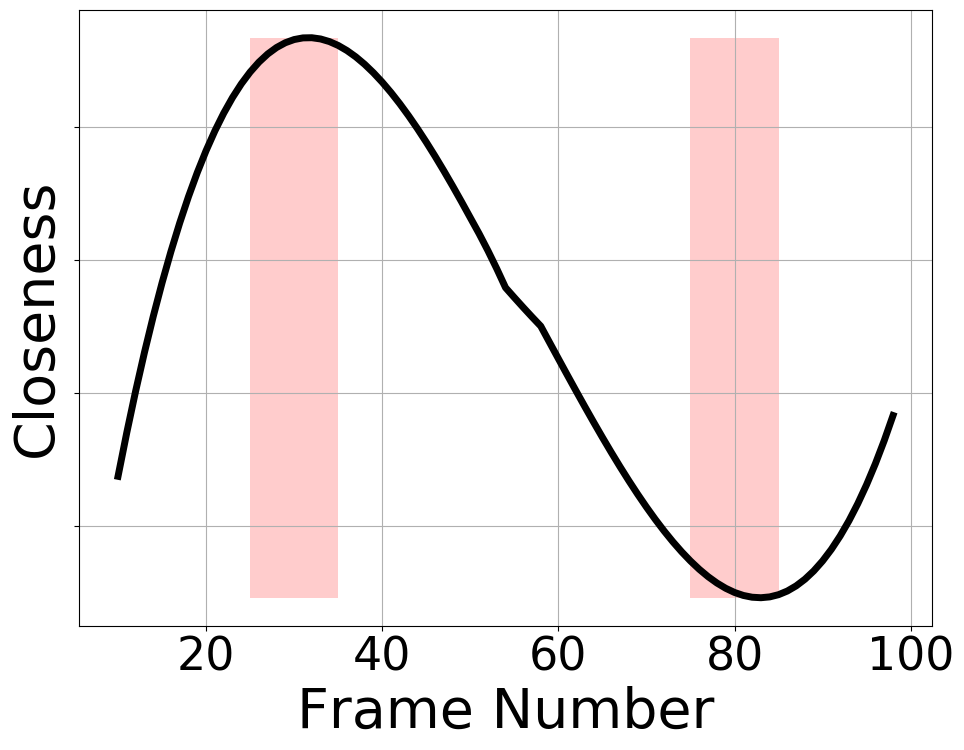}
    \caption{Extreme points for closeness centrality function for weaving vehicle.}
    \label{fig: Frame 3}
  \end{subfigure}
  \caption{\textbf{Measuring the Likelihood of Specific Styles with CMetric:} CMetric measures (degree and closeness centrality) the likelihood of the specific style of the ego-vehicle (grey with a blue glow) by computing the magnitude of the derivative of the centrality functions as well as the functions' extreme points. In Figure~\ref{fig: Frame 1}, the derivative of the degree centrality function is $0$ because the ego-vehicle does not observe any additional new neighbors (See Section~\ref{sec: centrality}) so the degree centrality is a constant function; therefore the vehicle is conservative. In Figure~\ref{fig: Frame 2}, the vehicle overspeeds, and consequently, the rate of observing new neighbors is high, which is reflected in the magnitude of the derivative of the degree centrality being positive. Finally, in Figure~\ref{fig: Frame 3}, the ego-vehicle demonstrates overtaking/sudden lane-changes and weaves through traffic. This is reflected in the magnitude of the slope and the location of extreme points, respectively, of the closeness centrality function. Our approach can classify these behaviors based on CMetric.}
  \label{fig: maneuver_behaviors}
  \vspace{-10pt}
\end{figure*}

Centrality functions~\cite{rodrigues2019network} are real-valued functions that characterize the behavior of vertices in a graph. Such functions can be defined as $\zeta:\mathcal{V}\longrightarrow \mathbb{R}$, where $\mathcal{V}$ denotes the set of vertices and $\mathbb{R}$ denotes a real number. The particular characteristic that is modeled through centrality functions depends on the type of graph. A few examples of centrality functions include closeness centrality, degree centrality, and eigenvector centrality. These functions have been used to identify influential personalities in social media networks~\cite{cen-socialmedia}, identify key infrastructure nodes on the internet~\cite{cen-internet}, rank web-pages in search engines~\cite{cen-pagerank}, and to discover the origin of epidemics~\cite{cen-epidemic}. In our approach, we show that centrality functions can measure the likelihood and intensity of different driver styles.



\begin{definition}
\textbf{Closeness Centrality:} In a connected traffic-graph at time $t$ with adjacency matrix $A_t$, let $\mc{D}_t(v_i,v_j)$ denote the minimum total edge cost to travel from vertex $i$ to vertex $j$, then the discrete closeness centrality measure for the $i^\textrm{th}$ vehicle at time $t$ is defined as,
\begin{equation}
    \zeta^i_c[t] = \frac{N-1}{\sum_{v_j\in \mathcal{V}(t)\setminus \{v_i\}} \mc{D}_t(v_i,v_j)},
    \label{eq: closeness}
\end{equation}
\end{definition}

The closeness centrality for a given vertex computes the reciprocal of the sum of the edge lengths of the shortest paths between the given vertex and all other vertices in the connected DGG. By definition, the higher the closeness centrality value, the more centrally the vertex is placed.

\begin{definition}
\textbf{Degree Centrality:} In a connected traffic-graph at time $t$ with adjacency matrix $A_t$, let $\mc{N}_i(t) = \{ v_j \in \mc{V}(t), \ A_t(i,j) \neq 0, \nu_j \leq \nu_i\}$ denote the set of vehicles in the neighborhood of the $i^\textrm{th}$ vehicle with radius $\mu$, then the discrete degree centrality function of the $i^\textrm{th}$ vehicle at time $t$ is defined as,
\begin{equation}
    \begin{aligned}
    \zeta^i_d[t] = \bigl | \{ v_j \in \mc{N}_i(t) \} \bigr | + \zeta^i_d[t-1] &\\
    \textrm{such that} \ (v_i,v_j) \not\in \mc{E(\tau)}, \tau = 0, \ldots, t-1&
    \end{aligned}
    \label{eq: degree}
\end{equation}
where $\vts{\cdot}$ denotes the cardinality of a set and $\nu_i, \nu_j$ denote the velocities of the $i^\textrm{th}$ and $j^\textrm{th}$ vehicles, respectively.
\end{definition}

The degree centrality at any time-step $t$ computes the cumulative number of edges between the given vertex and connected vertices in the traffic-graph over the past $t$ seconds, given that the velocity of the given vertex is higher than that of the connected vertices.


We exploit the analytical properties, including the derivatives and extreme values of these centrality functions, to develop our new metric in the following section. This metric computes the likelihood and intensity of different driving styles.
\section{Driving Behavior Classification using CMetric}
\label{sec: centrality}

We use centrality functions to develop a novel metric which answers two questions,
\begin{itemize}
    \item (Likelihood) At time $t$, how likely is it that a vehicle executes a specific style? 
    \item (Intensity) At time $t$, what is the intensity with which a vehicle executes a specific style?
\end{itemize}

The \textit{specific styles} can then be used to assign \textit{global behaviors} according to Table~\ref{tab: behaviors_centrality}. Our model (Figure~\ref{fig: overview}) is a computational model for behavior classification and consists of the following steps: 


\begin{enumerate}
    \item Obtain the positions of all vehicles using sensors deployed on the autonomous vehicle and form Dynamic Geometric Graphs (Section~\ref{subsec: DGG}).
    
    
    \item Compute the closeness and degree centrality function values using the definitions in Section~\ref{subsec: background_centrality}.
    
    \item Use the CMetric value to measure the likelihood and intensity of specific driving styles listed in Table~\ref{tab: behaviors_centrality}.
    
    
\end{enumerate}



\begin{algorithm}
    \SetKwInOut{Input}{Input}
    \SetKwInOut{Output}{Output}
\SetKwComment{Comment}{$\triangleright$\ }{}
\SetAlgoLined
\Input{$u = v_i \gets [x_i,y_i]^\top \ \forall v_i \in \mathcal{V}(t)$}
\Output{$\textrm{SLE}_k(t), \textrm{SIE}_k(t)$}
$t=0$\\
\For {each $u \in \mathcal{V}(t)$}{
\While{$t \leq T$}{ 
// Compute Centrality //\\
$     \zeta^i_c[t] = \frac{N-1}{\sum_{v_j\in \mathcal{V}(t)\setminus \{v_i\}} \mc{D}_t(v_i,v_j)}$ \\ 
$    \zeta^i_d[t] = \bigl | \{ v_j \in \mc{N}_i(t) \} \bigr | + \zeta^i_d[t-1], \ (v_i,v_j) \not\in \mc{E(\tau)}, \tau = 0, \ldots, t-1$ \\
$t \gets t+1$
 }
 // Compute CMetric //\\
Form $\cm$ using Definition~\ref{def: cmetric}\\
\For{$k=0,1$}{
// Compute Likelihood and Intensity // \\
$\textrm{SLE}_k(t) = \abs*{\frac{\partial \cm[k,:]}{\partial t}}$\\
$\textrm{SIE}_k(t) = \abs*{\frac{\partial^2 \cm[k,:]}{\partial t^2}}$
}
}
\caption{CMetric Measure outputs the Style Likelihood Estimate (SLE) and Style Intensity Estimate (SIE) for a vehicle, $u$, in a given time-period $\Delta t$. }
\end{algorithm}


In a given time-period $\Delta t$, we use the notion of CMetric to measure various driving styles. The CMetric is defined as:
\begin{definition}
\textbf{CMetric ($\mathcal{M}_{\Delta t}(u)$):} Given a time period $\Delta t$, the CMetric value for a particular vehicle, $u$, is a matrix $\cm \in \mathbb{R}^{2 \times \Delta t}$, where
$\cm = \left[
\begin{array}{c}
\zeta_c(u) \Tstrut \Bstrut \\
 \zeta_d(u) \Tstrut \Bstrut \\
\end{array}
\right]$.
\label{def: cmetric}
\end{definition}

\noindent where $\zeta_c(u)$ and $\zeta_d(u)$ are the closeness and degree centrality functions defined in Section~\ref{subsec: background_centrality}. We use the CMetric to measure the Style Likelihood Estimate (SLE) and Style Intensity Estimate (SIE) of driving behavior styles.

\begin{enumerate}
    \item The Style Likelihood Estimate (SLE) of a specific driving style is the probability of its occurrence and is measured by computing the magnitude of the row derivatives of $\cm$ with respect to time. Higher magnitudes of the row derivatives and the existence of local extreme points indicate higher likelihood. The SLE formula is given by,

\begin{equation}
    \textrm{SLE}_k(t) = \abs*{\frac{\partial \cm[k,:]}{\partial t}}
    \label{eq: SLE}
\end{equation}

    \item The Style Intensity Estimate (SIE) of a specific driving style is the severity with which the style is executed by a vehicle. Specific styles executed over shorter time-frames are considered more intense than styles executed over longer time-frames. SIE is computed by taking the second row derivatives of $\cm$ with respect to time and measuring the $\varepsilon$-sharpness of local extreme points. The SIE formula is given by,

\begin{equation}
    \textrm{SIE}_k(t) = \abs*{\frac{\partial^2 \cm[k,:]}{\partial t^2}}
    \label{eq: SIE}
\end{equation}

\end{enumerate}

\noindent The row index $k$ indicates the centrality function that is used. $k=0,1$ corresponds to the closeness and degree centrality, respectively. We can compute the time $t_\textrm{SLE}$ of maximum likelihood using 

\[t_\textrm{SLE} = \argmin_{t \in \Delta t}{\textrm{SLE}_k(t)}\]

\subsubsection{Overtaking/Sudden Lane-Changes}
Overtaking is the act of one vehicle going past another vehicle, traveling in the same or adjacent lane, in the same direction. From Definition~\ref{eq: closeness} in Section~\ref{subsec: background_centrality}, the value of the closeness centrality increases a vehicle moves towards the center and decreases as it moves away from the center. Therefore, the SLE of overtaking can be computed by measuring the rate of change of the closeness centrality. The maximum likelihood $\textrm{SLE}_\textrm{max}$ is,

\[\textrm{SLE}_{0,\textrm{max}} = \max_{t \in \Delta t}{\textrm{SLE}_0}(t).\]
\noindent The SIE of overtaking is computed by simply measuring SIE$_0$.

\subsubsection{Weaving} Weaving is the act of a vehicle shifting its position from a side lane towards the center, and vice-versa~\cite{farah2009passing}. In such a scenario, the closeness centrality function values oscillates between low values on the side lanes and high values towards the center. Mathematically, these oscillations in the closeness centrality values can be detected by finding the extreme values (points at which function has a local minimum or maximum) of the closeness centrality function. Mathematically, the points of local maximum or minimum can be found at those time instances when $\textrm{SLE}_0(t) = 0$. To differentiate from constant functions, we impose the condition that the $\varepsilon-$sharpness~\cite{dinh2017sharp} of the closeness centrality be non-zero:
\[\max_{t \in \mathcal{B}_\varepsilon(t^{*})} \textrm{SLE}_0(t) - \textrm{SLE}^{*}_0 \neq 0,\]
\noindent where $\mathcal{B}_\varepsilon (y) \in \mathbb{R}^d$ is the unit ball centered around a point $y$ with radius $\varepsilon$ in $d$ dimensions, and $t^{*} = \argmin_t{\textrm{SLE}_0}$. The $\textrm{SIE}_0(t)$ is computed by measuring the sharpness of the local minimum or maximum which is expressed using the $\varepsilon-$sharpness value.

\subsubsection{Overspeeding}

We use the degree centrality to classify overspeeding. The degree of $i^{\textrm{th}}$ vehicle, ($\theta_i \leq n$), at time $t$, can be computed from the diagonal elements of the Adjacency matrix $A_t$ (Section~\ref{subsec: DGG}). As $A_t$ is formed by adding rows and columns to $A_{t-1}$, the degree of each vehicle monotonically increases. Intuitively, an aggressively overspeeding vehicle will observe new neighbors (increasing degree) at a higher rate than neutral or conservative vehicles. Therefore, the likelihood of overspeeding can be measured by computing $\textrm{SLE}_1(t)$. Similar to overtaking, the maximum likelihood estimate is given by 
\[\textrm{SLE}_{1,\textrm{max}} = \max_{t  \in \Delta t}{\textrm{SLE}_1}(t)\]

Figures~\ref{fig: Frame 1} and~\ref{fig: Frame 2} visualizes how the degree centrality can distinguish between an overspeeding vehicle and a vehicle driving at a uniform speed.

\subsubsection{Conservative Vehicles}

Conservative vehicles, on the other hand, conform to a single lane~\cite{ahmed1999modeling} as much as possible, and drive at a uniform speed~\cite{sagberg2015review}, typically at or below the speed limit. The values of the closeness and degree centrality functions in the case of conservative vehicles thus, remain constant. More formally, the likelihood that a vehicle stays in a single lane during time-period $\Delta t$ is higher when,

\[\textrm{SLE}_0(t) = 0 \  \textrm{and} \ \max_{t \in \mathcal{B}_\varepsilon(t^{*})} \textrm{SLE}_0(t) - \textrm{SLE}^{*}_0 = 0,\]
and the likelihood that a vehicle will prefer to drive at uniform speed is higher when,
\[\textrm{SLE}_1(t) = 0 \  \textrm{and} \ \max_{t \in \mathcal{B}_\varepsilon(t^{*})} \textrm{SLE}_1(t) - \textrm{SLE}^{*}_1 = 0.\]

\noindent The more conservative a driver is, the higher their intensity value and can be similarly measured by computing the $\textrm{SIE}_k(t), k=0,1$.

In summary, the specific driving styles listed in Table~\ref{tab: behaviors_centrality} can be characterized by CMetric by exploiting its analytical properties, including the derivative, second derivative, extreme values, and the notion of $\varepsilon-$sharpness of the local minimum (or maximum) regions.

\begin{figure*}
\centering

 \includegraphics[width=\textwidth]{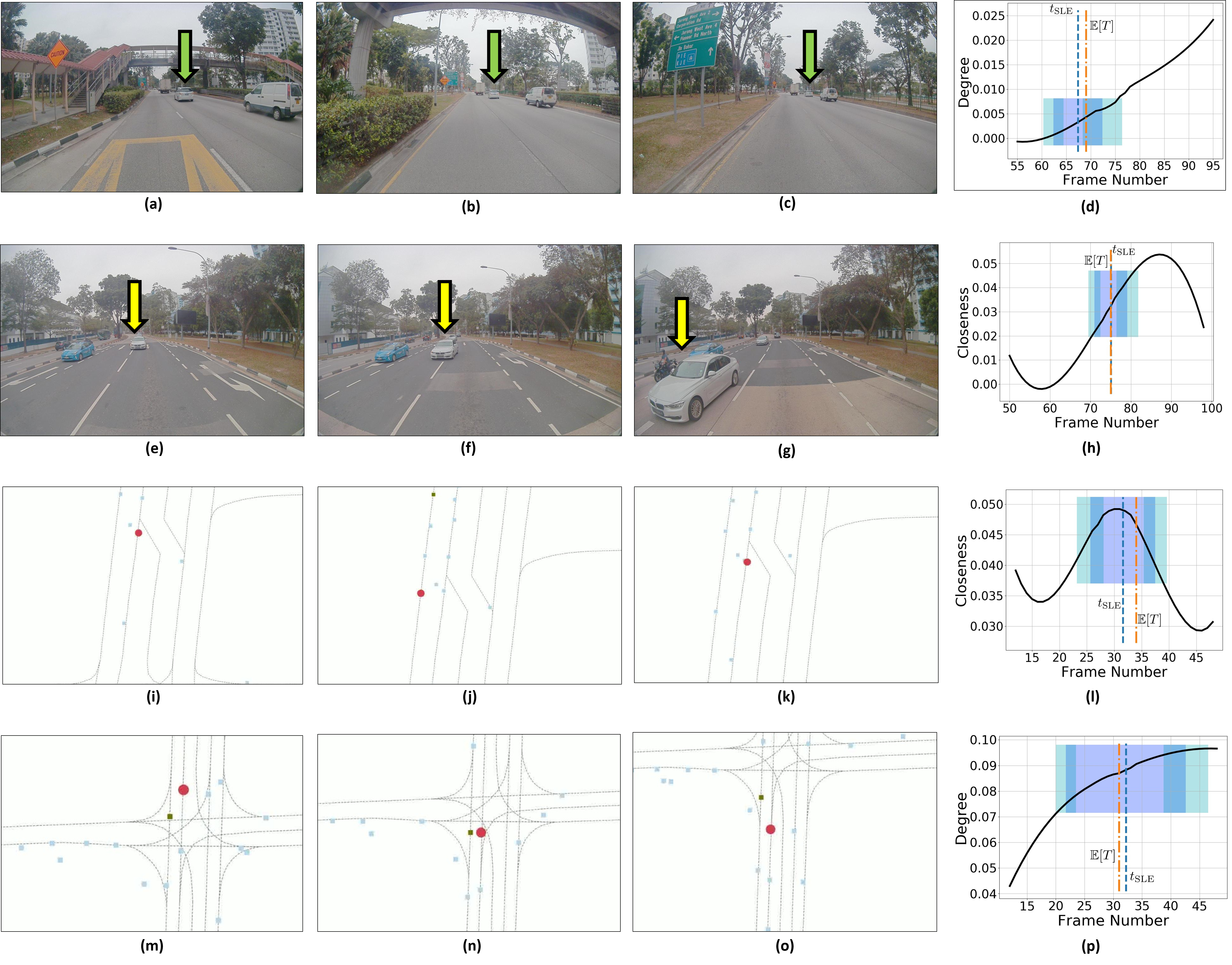}

\label{fig: deg_a}

 \caption{\textbf{Qualitative Analysis:} First and second rows are videos from the Singapore dataset, while the third and fourth rows correspond to videos from the Argoverse dataset. In each row, the first three figures demonstrate the trajectory a driving style executed by the ego-vehicle while the fourth figure shows the corresponding graph. The shaded colored regions overlaid on each graph are color heat maps that correspond to $\mathcal{P}(T)$. \textbf{Observation:} The expected time frame of a driving style reported by the participants of the user study matches that of the time of maximum likelihood computed by CMetric almost identically. }
  \label{fig: qualitative}
  \vspace{-15pt}
\end{figure*}

\section{Implementation and Evaluation}
\label{sec: experiments}

We first highlight the datasets used in our work, followed by the evaluation protocol. We report the results for metrics in Section~\ref{subsec:results}. We conclude the section with an analysis of the CMetric performance.

\subsection{Dataset}

One of the main issues in driving behavior research is the availability of large-scale open-source datasets for driving behaviors that specifically contain labels for aggressive and conservative vehicles. In light of these limitations, we create such datasets by both manually collecting traffic videos recorded in Singapore as well as by modifying existing large-scale autonomous driving datasets originally intended for trajectory prediction and tracking~\cite{Argoverse,lyft2019,ma2018trafficpredict}. In both cases, the behavior labels are annotated and verified via crowd-sourced annotators. We show samples of the dataset in our supplementary video.

\subsection{Evaluation Protocol}
As a driver's behavior changes continuously in traffic, an ideal evaluation protocol should predict the correct behavior \textit{during the time-period in which that behavior is observed}. Our proposed protocol, called the Time Deviation Error (TDE), measures the temporal difference between a human prediction and a model prediction. For example, if a vehicle executes a rash overtake at the $5^\textrm{th}$ frame and our model predicts the behavior at the $7^\textrm{th}$ frame, then the TDE$=0.067$ seconds assuming $30$ frames per second. A lower value for TDE indicates a more accurate behavior prediction model. The TDE is given by the following equation,

\begin{equation}
    \textrm{TDE}_\textrm{style} =   \abs*{\frac{t_\textrm{SLE} - \EX[T]}{f}} 
    \label{eq: TDE}
\end{equation}

\noindent where $\EX[T]$ denotes the expected time-stamp of an exhibited behavior in the ground-truth and $f$ is the frame rate of the video. $f=2$ Hz for the Singapore dataset and $f=10$ Hz for the U.S. dataset. In other words, the $\textrm{TDE}_\textrm{style}$ computes the time difference between the mean frame, $\EX[T]$, reported by the participants and the frame with the maximum likelihood, $t_\textrm{SLE}$, predicted by our approach. While $t_\textrm{SLE}$ is computed using $\argmax_{t \in \Delta t}{\textrm{SLE}(t)}$ as explained in Section~\ref{sec: centrality}, we discuss the computation of $\EX[T]$ in the technical report.

\begin{table}[t]
\centering
\caption{We report the Time Deviation Error (TDE) (in seconds) for the following driving styles: Overspeeding (OS), Overtaking (OT), Sudden Lane-Changes (SLC), and Weaving (W). The TDE indicates the absolute difference between the times taken by a human and our proposed CMetric to identify a driving style. Lower is better. $-$ indicates that the particular style was not observed in the ground truth.}
\centering
\resizebox{0.8\columnwidth}{!}{%
\begin{tabular}{lcccc} 
\toprule
 Dataset\Bstrut &  \multicolumn{4}{c}{Styles} \\
\cline{2-5}
& OS & OT & SLC & W\\
\midrule
Argoverse~\cite{Argoverse} \Tstrut & 0.25s & $-$  & 0.23s & $-$  \\
SG & 0.54s & 0.875s & 1.21s & 1.28s \\
\bottomrule
\end{tabular}
}
\label{tab:accuracy}
\vspace{-10pt}
\end{table}

\subsection{Results}
\label{subsec:results}
We report the Time Deviation Error (TDE), in seconds(s), for the following driving styles: Overspeeding (OS), Overtaking (OT), Sudden Lane-Changes (SLC), and Weaving (W), respectively, in Table~\ref{tab:accuracy}. We report the average TDE observed for each style across all videos in a dataset. For the Singapore dataset, we find that on average, the CMetric model automatically identifies an overspeeding agent half a second after a human prediction. Predictions for some behaviors in the Argoverse dataset could not be made as those behaviors were not observed in that dataset.


\subsection{Analysis of CMetric} 
\label{subsec: cmetric_analysis}
In Figure~\ref{fig: qualitative}, we show two sequences each from the Singapore and Argoverse datasets where we use CMetric to characterize overspeeding, overtaking, and sudden lane-changes using the degree and closeness centrality functions, respectively. Figures (a), (b), and (c) show the trajectory of an overspeeding vehicle during the $60^\textrm{th}$, $64^\textrm{th}$, and $69^\textrm{th}$ frames. In Figure (d), we show a monotonically rising curve for the degree function. From Section~\ref{sec: centrality}, the likelihood of overspeeding increases with increasing values of $\textrm{SLE}_1(t)$. $\textrm{SLE}_1(60) < \textrm{SLE}_1(64) < \textrm{SLE}_1(69)$ verifies the CMetric measure for overspeeding. Similarly, Figures (e), (f), and (g) show the trajectory of an overtaking vehicle during the $73^\textrm{rd}$, $75^\textrm{th}$, and $78^\textrm{th}$ frames. From Section~\ref{sec: centrality}, the likelihood of overtaking increases with increasing values of $\textrm{SLE}_0(t)$. In Figure (h), we show a monotonically rising curve for the closeness function between the $70^\textrm{th}$, and $80^\textrm{th}$ frames. In fact, we also observe extreme values of the closeness centrality function at the $57^\textrm{th}$, and $89^\textrm{th}$ frames. The CMetric measure for weaving (Section~\ref{sec: centrality}) indicates that the vehicle is also weaving through traffic at these time-instances.



\section{Conclusion, Limitations, and Future Work}

We present a new measure (CMetric) to classify driver behaviors using centrality functions. CMetric computes the likelihood of a vehicle executing a driving style along with the intensity used to execute the style.  Our approach is designed for realtime autonomous driving applications, where the trajectory of each vehicle or road-agent is extracted from a video. We represent traffic using dynamic geometric graph (DGG) and centrality functions, corresponding to closeness and degree, that are used to compute the CMetric. 

Currently, CMetric is limited to four driving styles, and we plan to investigate additional driving styles such as tailgating. Additionally, current evaluation was performed on straight roads and it would be interesting to test this approach in different settings such as intersections and traffic lights. As part of future work, we plan to use the CMetric measure for improving realtime planning and decision-making.

\section{Acknowledgements}

This work was supported in part by ARO Grants W911NF1910069 and W911NF1910315, Semiconductor Research Corporation (SRC), and Intel.






{\small
\bibliography{refs}
\bibliographystyle{IEEEtran}
}

\end{document}